\documentclass[runningheads]{llncs}
\usepackage[T1]{fontenc}
\usepackage{graphicx}
\usepackage{xcolor}
\newenvironment{DIFnomarkup}{}{}
\usepackage{booktabs}
\usepackage{hyperref}
\usepackage{color}
\usepackage[most]{tcolorbox}

\usepackage{amsmath,amsfonts,latexsym,amssymb,euscript,xr}
\usepackage{booktabs}
\usepackage[nodayofweek]{datetime}
\usepackage{float}
\usepackage[table]{xcolor}
\usepackage{multirow}
\usepackage{color,colortbl,tabularx}
\usepackage{amsmath,amssymb}
\usepackage{enumitem}

\usepackage{makecell}
\usepackage{verbatim}
\usepackage[english]{babel}
\usepackage[protrusion=true,expansion=true]{microtype}
\usepackage{amsmath,amsfonts}
\usepackage{pifont}
\usepackage{cleveref}
\usepackage{subcaption,graphicx}

\definecolor{LightBlue}{rgb}{0.88,0.9,0.9}

\begin{document}
\title{Novel Knowledge-Guided Generative Methods for Synthetic Transcriptomic Data}
%
%
\author{Francesca Pia Panaccione \inst{1}\orcidID{0009-0005-8007-963X} \and
 Sofia Mongardi\inst{1}\orcidID{0000-0002-5628-9101} \and
Marco Masseroli\inst{1}\orcidID{0000-0003-2574-1174}\and
Pietro Pinoli\inst{1}\orcidID{0000-0001-9786-2851}}
\authorrunning{F.P. Panaccione et al.}
%
\institute{DEIB - Dipartimento Elettronica, Informazione e Bioingegneria, Politecnico di Milano, Milan, Italy \\
\email{\{name.surname\}@polimi.it}}
\maketitle              
\begin{abstract}
As biomedical research increasingly relies on data-intensive tools, the quality and utility of datasets are critical. Challenges such as imbalances, biases, and ethical or legal constraints often limit access to high-quality data. Synthetic data generation can help overcome these limitations. Here, we present a comparative analysis of generative models for transcriptomic data, investigating strategies to incorporate prior biological knowledge via gene graphs. This ensures that synthetic data capture real-world gene patterns, maintaining their usefulness for downstream tasks. In particular, we introduce and benchmark three variants of the Generative Adversarial Network. Among the alternatives, MK-TGAN — an innovative multi-kernel, Graph Neural Network–based model — stands out for its performance in terms of both the realism and utility of the generated data. Unlike other methods, MK-TGAN leverages prior knowledge graphs by exploiting graph neural networks. Our results show that prior knowledge integration strategies improve performance, and that MK-TGAN consistently produces synthetic samples with superior realism and biological plausibility.
\end{abstract}

\section{\bf Introduction}
\label{sec:SCIENTIFIC-BACKGROUND}
Thanks to the advent of high-throughput sequencing technologies, data-intensive genomics has become increasingly relevant to biomedical research. The use of Machine Learning (ML) and Artificial Intelligence (AI) has grown exponentially, as reflected in PubMed, which listed 2,800 computationally grounded research articles in 2004, 7,600 in 2014, and 57,200 in 2024.

The success of data-driven approaches strongly depends on the availability of large, high-quality datasets. However, genomic data present several challenges, as they are intrinsically imbalanced—with underrepresented subgroups—and are subject to technical biases resulting from variations in collection and processing protocols. Moreover, their use and sharing are often constrained by ethical and legal considerations. In this scenario, the ability to generate synthetic data that accurately replicate the statistical properties of real genomic data could help address several challenges, such as rebalancing datasets (through data augmentation) and enabling their use without the legal and ethical limitations associated with data derived from patients~\cite{van2024}. Although synthetic data cannot be considered equivalent to real data—and any result derived from them (e.g., a trained model or a new hypothesis) must be biologically validated—they have the potential to accelerate many ML/AI-based research efforts that are affected by data scarcity. 

In recent years, deep learning-based generative methods have emerged as the most effective approach for this task. These methods typically begin with a simple and well-characterized latent distribution, such as a Gaussian or uniform distribution, from which random samples are drawn. Each sampled point is then transformed by a neural network that has been trained to map latent variables into synthetic data instances resembling real observations. Through this learned transformation, the model captures the statistical properties of the target distribution and generates new, realistic data accordingly.

Across multiple application domains, neural networks have been shown to benefit from the incorporation of domain knowledge that guides generation through explicit constraints, such as physical laws, clinical priors, or chemical properties. In the medical and biological domains, however, leveraging such information to ensure biologically plausible generation remains difficult~\cite{panaccione2026knowledge}, as biological knowledge is heterogeneous, incomplete, and shaped by inter-individual variability.

Recent work has explored integrating prior biological knowledge into generative models using graph-based representations, demonstrating that biological knowledge graphs can inform transcriptomic data generation. Although these approaches outperform unconditional models, the mechanisms by which biological graphs should be incorporated are not fully understood. Specifically, it is unclear whether implicit conditioning strategies—such as weight modulation or regularization—are sufficient to capture the structured dependencies inherent to gene expression data, or whether explicitly graph-aware architectural designs are required.

To address these open questions, we adopt biological knowledge graphs as a flexible representation of prior information and systematically investigate multiple strategies for integrating them into generative models based on the Generative Adversarial Network (GAN) framework.

We first introduce two graph-informed extensions of feed-forward GANs: Graph-Modulated GAN (GM-GAN) and Graph-Regularized GAN (GR-GAN). These models incorporate biological knowledge implicitly, either by modulating generator weights using gene–gene relationships or by regularizing parameters to align with the biological graph. While both approaches consistently improve data realism and downstream performance compared to unconditional models, they remain limited in their ability to capture the full variability of transcriptomic profiles. These limitations motivate the need for more expressive architectures that can more directly exploit structured biological relationships.

Building on these insights, we propose MK-TGAN (Multi-Kernel Transcriptomic Generative Adversarial Network), a novel framework that explicitly integrates biological knowledge through multiple graph neural network (GNN) kernels operating in parallel on a shared knowledge graph, thereby increasing representational capacity.

We evaluate all methods on breast cancer transcriptomic data using a biologically focused set of Epithelial-to-Mesenchymal Transition
(EMT)-related genes. This choice provides a controlled yet biologically meaningful setting, as EMT is a well-characterized process involving coordinated gene regulation and subtype-specific variability. This setup enables a focused assessment of how different knowledge-integration mechanisms affect the fidelity, diversity, and biological coherence of the generated samples. Empirically, MK-TGAN achieves state-of-the-art performance across precision, recall, and downstream biological tasks.

\subsection{Related Works} \label{RelWorks}
The generation of synthetic transcriptomic profiles has gained significant attention in recent years. In 2022, Viñas et al.~\cite{vinas2022adversarial} demonstrated the potential of deep learning, particularly Wasserstein GANs with Gradient Penalty (WGAN-GP)~\cite{arjovsky2017wassersteingan,gulrajani2017}, to generate realistic transcriptomic data for human and E. coli surpassing earlier methods based on simulation. Building on this, Lacan et al.~\cite{lacan2023gan} introduced an attention-based GAN to incorporate biological knowledge, focusing on genes with strong co-expression or protein-protein interaction links. However, their analysis suggested that performance gains were largely driven by architectural complexity rather than by the explicit integration of biological knowledge.
Other approaches have explored different strategies to inject domain information into transcriptomic generation. For instance, multimodal generative frameworks have been proposed to condition gene expression synthesis on heterogeneous inputs, such as histopathology images and clinical metadata~\cite{gemm-gan,pizurica2024digital}. Although these methods do not rely on graph priors, they demonstrate that external contextual information can effectively inform transcriptomic synthesis.

A related but distinct direction is represented by GOGGLE~\cite{liu2023goggle},  which marks one of the first attempts to explicitly align the architecture of a generative model with the relational structure of the data. By combining a Variational Autoencoder (VAE) with a message-passing neural network, GOGGLE embeds feature dependencies directly into the generative process, rather than treating them as external conditioning signals. While effective in low-dimensional tabular settings, this approach does not scale to transcriptomic data involving thousands of genes, limiting its applicability in high-dimensional biological domains.
Building upon this structural perspective, BioGAN~\cite{panaccione2025biogan} represents the first scalable generative framework explicitly designed for high-dimensional transcriptomic data. BioGAN integrates gene co-expression and regulatory networks directly into a GAN architecture through graph neural networks, thereby aligning the structure of the generator with known gene--gene dependencies. 

\section{\bf Data and Methods}
\label{sec:DATA-AND-METHODS}

This section describes the dataset used to train and test the models, and the architectures of the proposed GANs. We present three models that progressively incorporate prior biological knowledge in different ways: Graph-Modulated GAN (GM-GAN), Graph-Regularized GAN (GR-GAN), and Multi-Kernel Transcriptomic GAN (MK-TGAN). While all three share the same adversarial framework and training setup, they differ in the structure of the generator, which is the component responsible for integrating prior knowledge.

In our test, we consider the gene correlation matrix inferred from an external database as our prior knowledge. In particular, 
$A \in [-1, +1]^{n \times n}$, with $n$ corresponding to the number of genes, is the symmetric Pearson correlation matrix between every pair of genes.

\subsection{Model Overview}

\begin{figure*}[h]
    \centering
    \includegraphics[width=\textwidth]{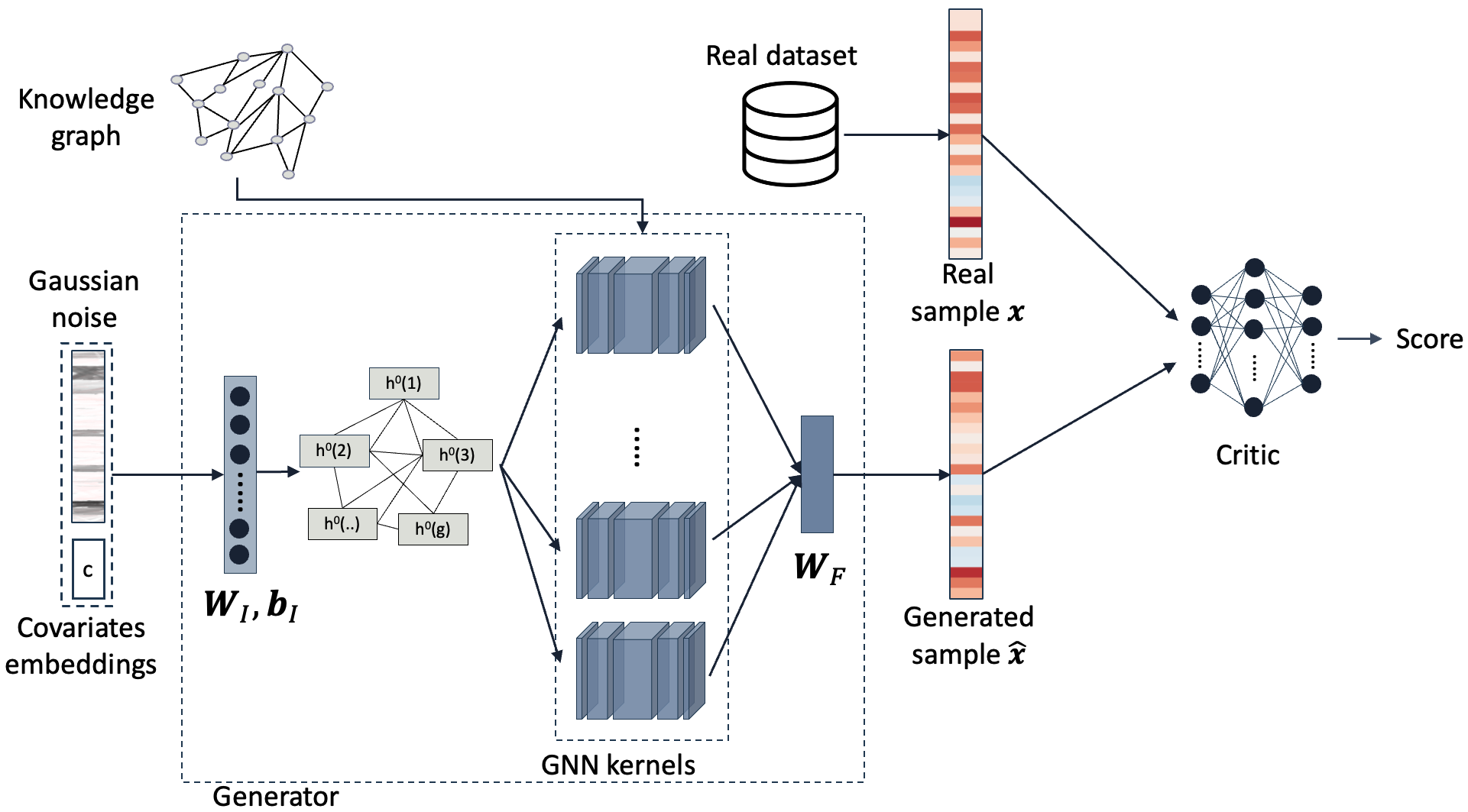}
    \caption{\textbf{MK-TGAN architecture for biologically informed gene expression synthesis.} An initial vector of noise is sampled from a \textbf{priori} distribution and passes through a fully connected layer that maps it to the nodes of a graph. Each node represents a single gene for which the expression value must be generated. The initial graph is processed in parallel by a set of GNN kernels, whose results are then combined using an element-wise multiplication with a weight matrix $W_F$. The generated sample is then passed to the critic that returns the final score. }
    \label{fig:graph_wgan_gp}
\end{figure*}

All models are based on an adversarial framework composed of two networks: a Generator $G$, which creates synthetic data from a noise vector $\mathbf{z}$ sampled from a known distribution $P_{\mathbf{z}}$, and a Discriminator $D$, which distinguishes real data from synthetic data. The two networks are trained in opposition to optimize the loss function:
$$\min_{G} \max_{D} \mathbb{E}_{x \sim p_{\text{data}}(x)}[\log D(x)] + \mathbb{E}_{z \sim p_{z}(z)}[\log(1 - D(G(z)))]. $$
Standard GANs are prone to training instability and mode collapse. This issue stems from the vanishing gradient problem, which occurs when the discriminator becomes so proficient that it can perfectly distinguish real samples from generated ones, leaving the generator with no meaningful gradient for improvement. Wasserstein GANs (WGANs) solve this problem by replacing the discriminator $D$ with a critic $C$ that measures the Wasserstein distance, yielding a continuous and informative gradient for the generator. The math behind the WGAN requires its critic $C$ to be 1-Lipschitz, meaning the norm of its gradient is at most 1 in all its domain. This can be achieved by adding to the optimization of the critic a term the penalizes its gradient:
{\small
\begin{equation}
    \min_{G} \max_{D} \left\{ 
    \mathbb{E}_{x \sim p_{\text{data}}(x)}[D(x)] 
    - \mathbb{E}_{z \sim p_{z}(z)}[D(G(z))] 
    - \lambda \mathbb{E}_{\hat{x} \sim p_{\hat{x}}}\left[
    \left( \| \nabla_{\hat{x}} D(\hat{x}) \|_{2} - 1 \right)^2 
    \right] 
    \right\}
\end{equation}
}
As illustrated in Figure~\ref{fig:graph_wgan_gp}, $G$ and $C$ can be further parametrized by a vector $c$ that represents the covariates on which the generation is conditioned. In all models, $\textbf{v} = [\textbf{z}, \textbf{c}]$ denotes the concatenated input vector combining the random noise and the conditioning variables, while $\hat{\textbf{x}} \in \mathbb{R}^{n}$ represents the generated gene expression profile for $n$ genes.

The remainder of this work compares a classical GAN and a standard WGAN-GP with three original WGAN-GP variants that incorporate prior knowledge to force the trained generator to follow biological properties.

\subsection{Graph-Modulated GAN (GM-GAN)}

The generator of the GM-GAN is a two layer dense feed forward network. Here, the prior biological knowledge is incorporated by modulating the learnable parameters of the last layer of the generator.

Let $\textbf{W} \in \mathbb{R}^{|genes|\times|genes|}$ denote the learnable weight matrix and $\textbf{b}\in \mathbb{R}^{|genes|}$ the bias term, a synthetic is progressively computated as follows:

\begin{align*}
&\textbf{x}_I = \text{LeakyReLU}(\textbf{v}\textbf{W}_I + \textbf{b}_I) \\
&\hat{\textbf{x}}  = \textbf{x}_I(\textbf{W}\odot \textbf{A}) + \textbf{b} \\
\end{align*}

where $\odot$ denotes the Hadamard (element-wise) product, $\textbf{W}_I$ and $\textbf{b}_I$ are the (learnable) parameters of the first layer.

By using $\textbf{A}$ in this way, the generator’s weights are scaled according to the strength of known biological relationships, promoting realistic dependencies among genes in the generated transcriptomic profiles.

\subsection{Graph-Regularized GAN (GR-GAN)}
The generator of the GR-GAN share the same topology of the one of GM-GAN. However, in this architecture the weighted prior knowledge matrix $\textbf{A}$ is not directly applied to the generator weights. Instead, it is incorporated indirectly by regularizing the learned weight structure, allowing the generator to remain flexible while still being guided by prior knowledge.
The generator mapping can be formalized as:
\begin{align*}
&\textbf{x}_I = \text{LeakyReLU}(\textbf{v}\textbf{W}_I + \textbf{b}_I) \\
&\hat{\textbf{x}}  = \textbf{x}_I\textbf{W} + \textbf{b} \\
\end{align*}
To integrate prior knowledge, the cost function (1) is augmented with the following two terms to be minimized:

\begin{equation*}
\lambda_1 \|\textbf{W} - \textbf{A}\|_2 + \lambda_2 \sum_{i,j} {ReLU(-(\textbf{A} \odot \textbf{W})_{i,j}})
\end{equation*}

The first term enforces structural similarity between the learned weight matrix $\textbf{W}$ and the prior knowledge matrix $\textbf{A}$, ensuring that the overall connectivity pattern of the generator reflects known gene-gene relationships.
 The elements $W_{ij}$ and $A_{ij}$ denote the entries of $\textbf{W}$ and $\textbf{A}$, respectively, representing the learned and prior interaction strength between genes $i$ and $j$.
Specifically, for each pair of genes $(i,j)$, the term $ReLU(-W_{ij}A_{ij})$ increases when $W_{ij}$ and $A_{ij}$ have opposite signs, thus discouraging learned relationships that contradict prior biological knowledge.
Compared to GM-GAN, which directly modulates the generator weights using $\textbf{A}$, GR-GAN allows the weights to remain fully learnable while being softly guided by the prior knowledge.

\subsection{Multi Kernel Transcriptomic GAN (MK-TGAN)}

MK-TGAN constitutes the most original contribution of this work. In contrast to the previous approaches, it introduces an architectural innovation specifically designed to model the graph-structured nature of transcriptomic data while adapting to the topology defined by the prior biological knowledge.

Similar to the previous approaches, the distinctive feature of MK-TGAN resides in its generator architecture, which employs multiple GNN kernels connected to non-learnable graphs, where genes are represented as nodes and edge weights reflect prior biological knowledge, to guide the generation of synthetic data. In each kernel and at each layer, each gene is represented as a multidimensional embedding. Graph convolutional functions transform the representation of each node by combining its current embedding with those of its neighbors, weighted by the strength of their connections. 

The set of initial set of embeddings $\textbf{h}^0$ is computed as:
\begin{equation}
\textbf{h}^{0} = ReLU(\textbf{W}_{I}\textbf{v} + \textbf{b}_I)
\end{equation}
where $\textbf{W}_I \in \mathbb{R}^{|genes|\times |v|}$ and $\textbf{b}_I \in \mathbb{R}^{len(\textbf{v})}$ are two learnable tensors.

The generator employs multiple parallel GNN kernels, each consisting of several graph convolutional layers. Each kernel processes and modifies an independent embedding representation of each node.
The $l$-th layer of the $k$-th kernel generates the embedding of the $i$-th node $h^{l}_{i, k}$ as follows: 
\begin{equation}
\textbf{h}_{i,k}^{l} = ReLU \Big( \textbf{h}_{i,k}^{l-1}\textbf{W}_{1,l,k} 
+ \frac{1}{|N(i)|}  \sum_{ j \in N(i)} w_{j,i} \textbf{h}_{j,k}^{l-1} \textbf{W}_{2,l,k} 
  \big) \end{equation}

where $N(i)$ represents the set of neighbors of the node $i$, $w_{j,i}$ the weight of the association between $i$ and $j$, and $\textbf{W}_{1,l,k}$, $\textbf{W}_{2,l,k} \in \mathbb{R}^{d(l-1) \times d(l)}$ are two learnable matrices such that $d(l-1)$ and $d(l)$ are the dimensions of the node embeddings of the layers $l-1$ and $l$ 
respectively.

Each kernel is designed so that the last embeddings have a length of 1.
 The generated value $\hat{\textbf{x}}_i$, i.e., the expression of the gene $i$, is computed as: {\small
\begin{equation}
\hat{\textbf{x}}_i = \sum_{k=1}^{K} h_{i,k}^L\textbf{W}_F[i,k]
\end{equation}
}

where $\textbf{W}_F \in \mathbb{R}^{|\text{genes}| \times K}$ is a learnable weight matrix used to combine the outputs from the $K$ different kernels. This formulation enables the model to fuse information from multiple kernel to compute the final generated expression profile $\hat{\textbf{x}} \in \mathbb{R}^{|\text{genes}|} $.

\paragraph{Architectural novelty with respect to the state of the art.}
Compared to the graph-based transcriptomic generator proposed in~\cite{panaccione2025biogan} (Section~\ref{RelWorks}), MK-TGAN introduces a substantially more expressive mechanism for incorporating biological priors into the generative process. The method in~\cite{panaccione2025biogan} relies on a single learnable message-passing pathway operating over a fixed, non-learnable biological graph, thereby compressing all relational information into a single embedding space.

In contrast, MK-TGAN adopts a multi-kernel formulation where multiple independently parameterized GNN kernels operate in parallel on the same biological prior. By capturing complementary and heterogeneous biological relationships and adaptively combining them through a learnable matrix, the model achieves increased representational capacity and a more modular architecture. 

\subsection{Data}
\paragraph{Transcriptomic dataset.}
We trained and evaluated our model using a bulk RNA-seq dataset retrieved from The Cancer Genome Atlas (TCGA)~\cite{tcga} for Breast Cancer. The dataset comprises 1,127 samples annotated with tumor subtypes (Basal, Her2, LumA, and LumB). Gene expression values are provided in FPKM units and standardized (zero mean and unit variance) prior to model training. We focused on 197 genes associated with the Epithelial-to-Mesenchymal Transition (EMT) process, retrieved from MSigDB\footnote{\url{https://www.gsea-msigdb.org/gsea/msigdb}}.
\paragraph{Construction of the biological prior.}
To incorporate external biological structure, we leveraged 396 healthy breast tissue samples from the Genotype-Tissue Expression (GTEx) project~\cite{gtex2020tissue}. Using these samples, we constructed a co-expression–based knowledge graph by computing pairwise Pearson correlation coefficients between all EMT genes across samples. This graph serves as a biologically grounded prior used to guide the generative process and subsequent evaluation.
\paragraph{Graph representation.}
The resulting prior knowledge graph is encoded as a dense, symmetric gene–gene correlation matrix $A \in \mathbb{R}^{n \times n}$, with $n = 197$. The matrix spans the range $[-0.741,0.976]$ and contains no exact zeros in the off-diagonal entries. Both positive and negative correlations are retained, with negative values accounting for 38.6\% of all off-diagonal elements. To preserve the full structure of the co-expression signal, no post-hoc preprocessing steps—such as thresholding, top-$k$ filtering, or sign-based pruning—are applied.
\section{Experiments and Evaluation}

We evaluated all models using a unified experimental protocol designed to facilitate fair comparison and straightforward reproducibility. The gene–gene co-expression graph derived from GTEx data was used as domain knowledge for all models incorporating biological priors and was kept identical across all experiments.

The TCGA breast cancer dataset was stratified by molecular subtype and split into 80\% training and 20\% held-out test sets. All generative models were trained exclusively on the training split, with no access to test samples during training. Molecular subtype labels were used as conditioning information and implemented via learnable embedding layers, which were concatenated to the inputs of both the generator and the critic. All models shared the same critic—a three-layer fully connected network. For non-graph-based baselines, the generator consisted of a three-stage fully connected feed-forward architecture. In contrast, for graph-based models, the fully connected generator layers were replaced by graph neural network modules, in which latent features are transformed through message passing over the biological graph.

All methods were implemented in PyTorch and trained using the conditional WGAN-GP framework, employing the Wasserstein loss with gradient penalty ($\lambda = 100$) to enforce the 1-Lipschitz constraint. Models were trained for 600 epochs with a latent space of dimension 128, followed by a final projection layer matching the number of genes to be generated. Optimization was performed using RMSprop with learning rate $\text{lr}=5\times10^{-4}$. The critic was updated twice for every generator update, and a step-based learning rate scheduler reduced the learning rate by a factor of 0.5 every 150 epochs. Linear layers were initialized using Xavier initialization, while embedding layers were initialized from a zero-mean normal distribution.

In MK-TGAN, the generator incorporates $K=4$ parallel GNN kernels operating on the same biological graph. All experiments were run on a single NVIDIA A2 GPU (Ampere, 16,GB), with similar training times observed across all models. MK-TGAN’s multi-kernel architecture did not introduce a significant increase in training time despite its additional complexity, as the parallel kernels operate on relatively small graphs and share a common forward-pass structure.

For evaluation, each trained generator produced synthetic samples conditioned on the subtype labels of the held-out test set. This procedure was repeated for ten independent sampling runs. Synthetic samples were evaluated against the corresponding real test data using the same test split for all models and runs.

To assess the impact of biological prior integration, we compared our methods against multiple baselines. These include two models without prior biological knowledge (GAN and WGAN-GP) and a graph-informed generative model inspired by the BioGAN framework for high-dimensional transcriptomic data generation. To ensure maximal comparability, the BioGAN-inspired model was implemented using the same training protocol and architectural components as in our experiments, while preserving its original single-stream graph-based message-passing mechanism.

\subsection{Evaluation Metrics}

Model performance was assessed using two complementary classes of metrics: unsupervised and supervised.

\textit{Unsupervised metrics} quantify the statistical similarity between real and generated samples, providing insight into distributional fidelity. \textit{Supervised metrics} evaluate both the practical utility of synthetic samples and their indistinguishability from real data by means of standard machine learning classifiers trained on labeled data.

\subsubsection{Unsupervised Metrics}

\paragraph{Precision and Recall}
Precision and recall \cite{prec_recall} are metrics used to assess the quality and relevance of the generated data with respect to the real data. Given the real dataset $\textbf{X} =\{ x_1, \ldots, x_n\}$ and the generated dataset $\hat{\textbf{X}} = \{\hat{x}_1, \ldots, \hat{x}_m \}$, the first step is to compute the manifolds for the real and generated samples $H^t_\textbf{X}$ and $H^t_{\hat{\textbf{X}}}$, respectively. This is done by forming a sphere around each point of the two datasets with a radius corresponding to the euclidean distance to \textit{t}-th closest neighbor within the same dataset. A binary function $f(h, H)$ is then defined to evaluate whether the sample $h$ falls within the manifold $H$.
Precision and recall can be computed accordingly:

{\small
\begin{equation}
\text{precision}_t(\textbf{X}, \hat{\textbf{X}}) = \frac{1}{|\hat{\textbf{X}}|} \sum_{\hat{\textbf{x}} \in \hat{\textbf{X}}} f(\hat{\textbf{x}}, H^t_\textbf{X}),
\end{equation}
\begin{equation}
\text{recall}_t(\textbf{X}, \hat{\textbf{X}}) = \frac{1}{|\textbf{X}|} \sum_{\textbf{x} \in \textbf{X}} f(\textbf{x}, H^t_{\hat{\textbf{X}}}).
\end{equation}
}
 The precision is defined as the percentage of generated samples falling in the manifold of the real data, and conversely the recall is the percentage of true samples falling in the manifold estimated from the generated samples. We computed precision and recall considering the 10-th nearest neighbors.

\paragraph{Correlation Coefficient}
We used the similarity coefficient from \cite{lacan2023gan} to evaluate how well synthetic data maintain gene-gene relationships. This method, grounded in the Pearson correlation coefficient, measures similarity by comparing pairwise gene relationships between real and generated data.

Given two symmetric $n\times n$ matrices $\textbf{Y}$ and $\hat{\textbf{Y}}$ representing the pairwise distances between all features in the real and generated datasets, respectively, the similarity coefficient is defined as:  

{\small
\begin{equation}
c(\textbf{Y}, \hat{\textbf{Y}}) = \sum_{i=1}^{n-1} \sum_{j=i+1}^{n} 
\frac{(\textbf{Y}_{i,j} - l(\textbf{Y}))}{r(\textbf{Y})} \cdot 
\frac{(\hat{\textbf{Y}}_{i,j} - l(\hat{\textbf{Y}}))}{r(\hat{\textbf{Y}})}
\end{equation}
}

where for the $n \times n$ matrix $\textbf{Y}$,  $l(\textbf{Y})$ and $r(\textbf{Y})$ denote the mean and standard deviation of the pairwise values in matrix $\textbf{Y}$, and are defined as:

{\small
\begin{equation}
l(\textbf{Y}) = \frac{2}{n(n-1)} \sum_{i=1}^{n-1} \sum_{j=i+1}^{n} \textbf{Y}_{i,j}
\end{equation}
}

and
{\small
\begin{equation}
r(\textbf{Y}) = \sqrt{\frac{2}{n(n-1)} \sum_{i=1}^{n-1} \sum_{j=i+1}^{n} (\textbf{Y}_{i,j} - l(\textbf{Y}))^2}
\end{equation}
}

The same quantities are also derived for $\hat{\textbf{Y}}$.





\subsubsection{Supervised Metrics}
\paragraph{Utility (Train on Synthetic, Test on Real).}
To assess the utility of synthetic data, we adopt a Train on Synthetic, Test on Real (TSTR) protocol, in which classification models are trained exclusively on generated samples and evaluated on the held-out test set of real data. The objective is to predict breast cancer subtypes and evaluate how well synthetic data support downstream predictive tasks. We consider both a linear classifier, Logistic Regression (LR), and a non-linear neural classifier, a Multi-Layer Perceptron (MLP), in order to capture different modeling capacities. LR is trained with a maximum of 500 iterations, while the MLP employs two hidden layers of 128 units each and is trained for up to 300 iterations. Performance is reported using Balanced Accuracy and F1-score to account for class imbalance.

\paragraph{Detectability.}
Detectability measures the ability of a classifier to distinguish real samples from synthetic ones. For this task, the real test dataset and the corresponding generated samples are combined into a single dataset, resulting in a binary classification problem. Synthetic samples are generated in the same number as real samples and are associated with the same class labels used during generation; for the detectability task, samples are then labeled as synthetic ($1$) or real ($0$). The resulting dataset is subsequently split into 80\% train and 20\% test sets, and classification performance is evaluated on the held-out test split.

The same family of classifiers used in the utility evaluation is employed, with hyperparameters adapted to the binary setting. In particular, LR is trained for up to 1000 iterations, while MLP uses a single hidden layer of 100 units. Lower classification performance indicates higher similarity between real and synthetic data. Results are reported in terms of Accuracy and F1-score.




\section{\bf Results}
\label{sec:RESULTS}
Table~\ref{tab:results} reports the comparative evaluation of the generated samples across the considered methods.

Overall, MK-TGAN produces the highest-quality samples according to both unsupervised and supervised evaluations. In particular, MK-TGAN achieves the best recall (Recall$_{10}=0.784$) and the highest correlation score (0.914), while maintaining very high precision (0.960). The multi-kernel architecture appears to enhance the diversity of generated samples, resulting in improved coverage of the training distribution. By contrast, methods that achieve high precision but low recall—most notably BioGAN in this setting (Precision$_{10}=0.837$, Recall$_{10}=0.381$)—tend to accurately reproduce a limited subset of the real data, while failing to capture its broader variability. This behavior highlights a clear precision–recall trade-off: high precision alone reflects faithful reproduction of specific patterns, whereas high recall is required to cover a larger portion of the data manifold. MK-TGAN effectively balances this trade-off, likely due to its multi-kernel formulation, which enables the model to capture diverse relational patterns that may be compressed or lost in single-stream graph neural network architectures.

Detectability results further support this interpretation. MK-TGAN yields markedly lower detectability scores (e.g., MLP accuracy $=0.760$), indicating that generated samples are more similar to real data and therefore harder to distinguish. In contrast, several competing methods (GAN, WGAN-GP, and BioGAN) exhibit near-perfect detectability (accuracy $\approx 1.0$), consistent with the observation that these models generate a restricted or imbalanced coverage of the data distribution. GR-GAN achieves intermediate detectability values (LR accuracy $=0.923$), reflecting its softer, regularization-based integration of prior biological knowledge.
However, detectability remains a critical measure for evaluating generative performance. Even for MK-TGAN, the observed MLP accuracy suggests that the generated samples, while more realistic than those of other methods, still do not fully capture the diversity of the real dataset.

In the supervised utility evaluation (TSTR), all models incorporating prior knowledge—including BioGAN—outperform the baselines without prior knowledge, supporting the hypothesis that biological constraints improve the usefulness of synthetic data in downstream tasks. Notably, MK-TGAN consistently achieves the best downstream performance (e.g., LR accuracy $=0.805$, MLP accuracy $=0.791$), confirming that increased coverage and variability translate into more informative synthetic samples for predictive applications. These results indicate that structured biological priors constrain generation toward meaningful regions of the data space enhancing the biological coherence.


Finally, the proposed architecture is also more parameter-efficient. For the tested dataset, MK-TGAN uses fewer than 19,000 learnable parameters, compared to over 100,000 in the other models, which rely on fully connected layers. This substantial reduction suggests that the architecture may be less prone to overfitting and could be trained effectively with fewer data samples.

In summary, the experimental results indicate that (i) incorporating prior biological knowledge improves both the realism and utility of synthetic transcriptomic data compared to unconstrained baselines and implicit regularization strategies that do not rely on graph-structured generation (e.g., GR-GAN and GM-GAN); (ii) single-stream graph-informed generators can achieve high precision but may suffer from limited recall and reduced variability; and (iii) the multi-kernel MK-TGAN formulation provides a more balanced trade-off, increasing diversity (recall) while preserving high fidelity (precision).

\begin{DIFnomarkup}
\begin{table}[H]
\caption{Performance metrics for generated samples of genes involved in the EMT pathway, using TCGA Breast Cancer data as input. Results are shown for two baseline methods without prior knowledge (GAN and WPGAN), a state-of-the-art graph-based transcriptomic generator (BioGAN), and three proposed prior-informed methods (GM-GAN, GR-GAN, and MK-TGAN). Mean and standard deviation across 10 generation runs are reported. Bold indicates the best performance.}
\label{tab:results}
\centering
\small
\renewcommand{\arraystretch}{1.58}
\resizebox{\textwidth}{!}{%
\begin{tabular}{|c|c|c|c|c|c|c|}
\hline
\multirow{2}{*}{Metric} & \multicolumn{3}{c|}{State-of-the-art approaches} & \multicolumn{3}{c|}{Proposed approaches} \\ \cline{2-7}
 & GAN & WGAN-GP & BioGAN & GM-GAN & GR-GAN & MK-TGAN \\ \hline
\multicolumn{7}{|c|}{Unsupervised} \\ \hline
Precision$_{10}$ $\uparrow$ & 0.793 (0.020) & 0.696 (0.014) & 0.837 (0.019) & \textbf{0.967 (0.008)} & 0.794 (0.029) & 0.960 (0.015) \\
Recall$_{10}$ $\uparrow$    & 0.546 (0.026) & 0.668 (0.028) & 0.381 (0.036) & 0.300 (0.025) & 0.707 (0.027) & \textbf{0.784 (0.020)} \\
Correlation $\uparrow$      & 0.591 (0.013) & 0.696 (0.014) & 0.801 (0.015) & 0.813 (0.023) & 0.848 (0.011) & \textbf{0.914 (0.008)} \\ \hline
\multicolumn{7}{|c|}{Detectability} \\ \hline
LR (acc.) $\downarrow$ & 1.000 (0.000) & 1.000 (0.000) & 0.968 (0.001)& 1.000 (0.000) & 0.923 (0.019) & \textbf{0.624 (0.024)} \\
LR (F1) $\downarrow$   & 1.000 (0.000) & 1.000 (0.000) & 0.968 (0.001) & 1.000 (0.000) & 0.923 (0.019) & \textbf{0.622 (0.024)} \\
MLP (acc.) $\downarrow$& 1.000 (0.000) & 1.000 (0.000) & 1.000 (0.000) & 1.000 (0.000) & 1.000 (0.000) & \textbf{0.760 (0.013)} \\
MLP (F1) $\downarrow$  & 1.000 (0.000) & 1.000 (0.000) & 1.000 (0.000) & 1.000 (0.000) & 1.000 (0.000) & \textbf{0.759 (0.046)} \\ \hline
\multicolumn{7}{|c|}{Utility} \\ \hline
LR (acc.) $\uparrow$ & 0.602 (0.042) & 0.664 (0.024) & 0.725 (0.024) & 0.665 (0.027) & 0.731 (0.015) & \textbf{0.805 (0.028)} \\
LR (F1) $\uparrow$    & 0.570 (0.039) & 0.597 (0.027) & 0.718 (0.020) & 0.684 (0.023) & 0.733 (0.015) & \textbf{0.800 (0.022)} \\
MLP (acc.) $\uparrow$ & 0.630 (0.032) & 0.671 (0.021) & 0.730 (0.024) & 0.715 (0.015) & 0.714 (0.021) & \textbf{0.791 (0.022)} \\
MLP (F1) $\uparrow$   & 0.616 (0.034) & 0.647 (0.024) & 0.721 (0.020) & 0.714 (0.016) & 0.714 (0.019) & \textbf{0.780 (0.015)} \\ \hline
\end{tabular}}
\end{table}
\end{DIFnomarkup}

\section{Conclusion}

In this work, we address the generation of synthetic transcriptomic data as a strategy to mitigate dataset imbalance, bias, and data-sharing constraints, with the goal of improving both realism and downstream utility in breast cancer subtype prediction. Our results demonstrate that incorporating prior biological knowledge substantially enhances the statistical fidelity of generated samples and their effectiveness in predictive tasks.

We systematically compare multiple strategies for integrating biological priors and introduce MK-TGAN, a novel generative framework tailored to transcriptomic data. By explicitly leveraging gene–gene relationships through graph neural networks within the generator, MK-TGAN consistently outperforms baseline models as well as GM-GAN and GR-GAN, which represent softer, implicit forms of prior-knowledge integration.

A limitation of this study is that experimental validation is restricted to breast cancer transcriptomic data and to a biologically focused set of EMT-related genes. While this controlled setting facilitates a clear and interpretable analysis of different knowledge-integration mechanisms, it may not fully reflect the heterogeneity, scale, and noise characteristics of whole-transcriptome data or of other disease contexts.

Despite this limitation, the proposed architecture and learning strategy are not inherently tied to a specific gene set or disease and can, in principle, be extended to broader transcriptomic domains. Future work will therefore focus on evaluating MK-TGAN on larger gene panels and whole-transcriptome datasets, spanning multiple cancer types and biological processes, with the aim of further enhancing robustness and generalizability.

Finally, the residual distinguishability of the generated samples underscores the inherent difficulty of fully capturing transcriptomic complexity using GAN-based approaches. Addressing these limitations through the incorporation of richer biological priors—such as regulatory, pathway, and protein–protein interaction networks—and the adoption of more expressive graph-based architectures may enable more faithful in silico simulations of gene regulatory dynamics and disease mechanisms.

%
%
\bibliographystyle{splncs04}
\bibliography{bibliography_CIBB_file}
%




\end{document}